\documentclass[10pt,twocolumn,letterpaper]{article}
\usepackage[T1]{fontenc}
\usepackage{newtxtext,newtxmath}
\usepackage[letterpaper,textwidth=6.875in,textheight=8.875in,top=1in]{geometry}
\usepackage{microtype}
\usepackage{amsmath}
\usepackage{graphicx,booktabs,multirow,xcolor}
\usepackage[font=small,labelfont=bf,labelsep=period]{caption}
\usepackage[numbers,sort&compress]{natbib}
\usepackage{stfloats}
\usepackage{flushend}
\definecolor{linkblue}{RGB}{27,84,137}
\usepackage[colorlinks,allcolors=linkblue]{hyperref}

\newcommand{\papertitle}{Improve How You Improve: Recursive Policy Self-Improvement for Zero-Shot Open-Vocabulary Multi-Object Navigation}
\title{\bfseries\papertitle}
\author{
Yang Li\textsuperscript{1} \quad
Ziju Han\textsuperscript{1} \quad
Aming WU\textsuperscript{2} \quad
Zihao Zhang\textsuperscript{1}
\\[0.25em]
Sijia Zhang\textsuperscript{1} \quad
Yahong Han\textsuperscript{1}\thanks{Corresponding author.} \quad
Qinghua Hu\textsuperscript{1}
\\[0.6em]
\normalsize\textsuperscript{1}\,School of Artificial Intelligence, Tianjin University, China
\\
\normalsize\textsuperscript{2}\,School of Computer Science and Information Engineering, Hefei University of Technology, China
\\[0.45em]
\texttt{\small\{liyang1389, 3024244296, zhangzihao2490, hanziju, yahong\}@tju.edu.cn}
\\
\texttt{\small amwu@hfut.edu.cn \quad huqinghua@tju.edu.cn}
}
\date{}
\hypersetup{
  pdftitle={\papertitle},
  pdfauthor={Yang Li, Ziju Han, Aming WU, Zihao Zhang, Sijia Zhang, Yahong Han, Qinghua Hu}
}

\begin{document}
\maketitle

\begin{abstract}

To advance the development of embodied general navigation, zero-shot open-vocabulary multi-object navigation has attracted increasing attention, aiming to enable robots to efficiently search multiple objects in unseen environments without training. 
For this task, the key lies in improving navigation policies from earlier searches to locate subsequent goals more efficiently. Recent methods automate this process through trajectory reflection and rule revision, enabling policy self-improvement.
However, the mechanisms for diagnosing problems and generating revisions are typically designed by humans and remain fixed across iterations. Consequently, the system’s capacity for continued self-improvement remains constrained, even as its navigation policy evolves.
To this end, we propose \textbf{RPSI-Nav}, a
\textbf{R}ecursive \textbf{P}olicy \textbf{S}elf-\textbf{I}mprovement
framework for zero-shot multi-object navigation that revises both
the navigation program and its revision mechanism.
RPSI-Nav diagnoses search behavior, requests diagnostic views
when existing observations cannot distinguish competing causes,
and generates targeted navigation revisions.
Feedback from subsequent goal searches is linked to the rules
that produced these revisions, guiding changes to diagnosis,
observation acquisition, and revision planning.
The updated improver governs subsequent navigation improvements, forming a recursive loop in which RPSI-Nav revises its own improvement process.
Across goals, the robot conditionally inherits accepted program updates and spatial memory, enabling search experience to improve both future navigation and adaptation. Benchmark experiments and qualitative real-robot evaluations validate RPSI-Nav’s effectiveness.

\end{abstract}

\section{Introduction}
\label{sec:introduction}
Zero-Shot Object Navigation (ZSON) is a challenging task in Embodied AI, requiring an agent to navigate to target objects in unfamiliar
environments without training. Unlike conventional object navigation methods that focus on a single target, we consider zero-shot open-vocabulary multi-object
navigation, where a robot must search for and locate multiple
language-specified targets in succession 
within the same environment~\citep{onemap,wtrp_searcher}. 
In this setting, agents can use experience from earlier searches
to refine their navigation strategies and locate subsequent
goals more efficiently. When agents autonomously reflect on
past searches and revise their decision rules, this process
constitutes policy self-improvement.
Related work ~\cite{evolvenav,retronav,slow4fast} explores such self-improvement across navigation settings. 

\begin{figure*}
    \centering
    \includegraphics[width=\textwidth]{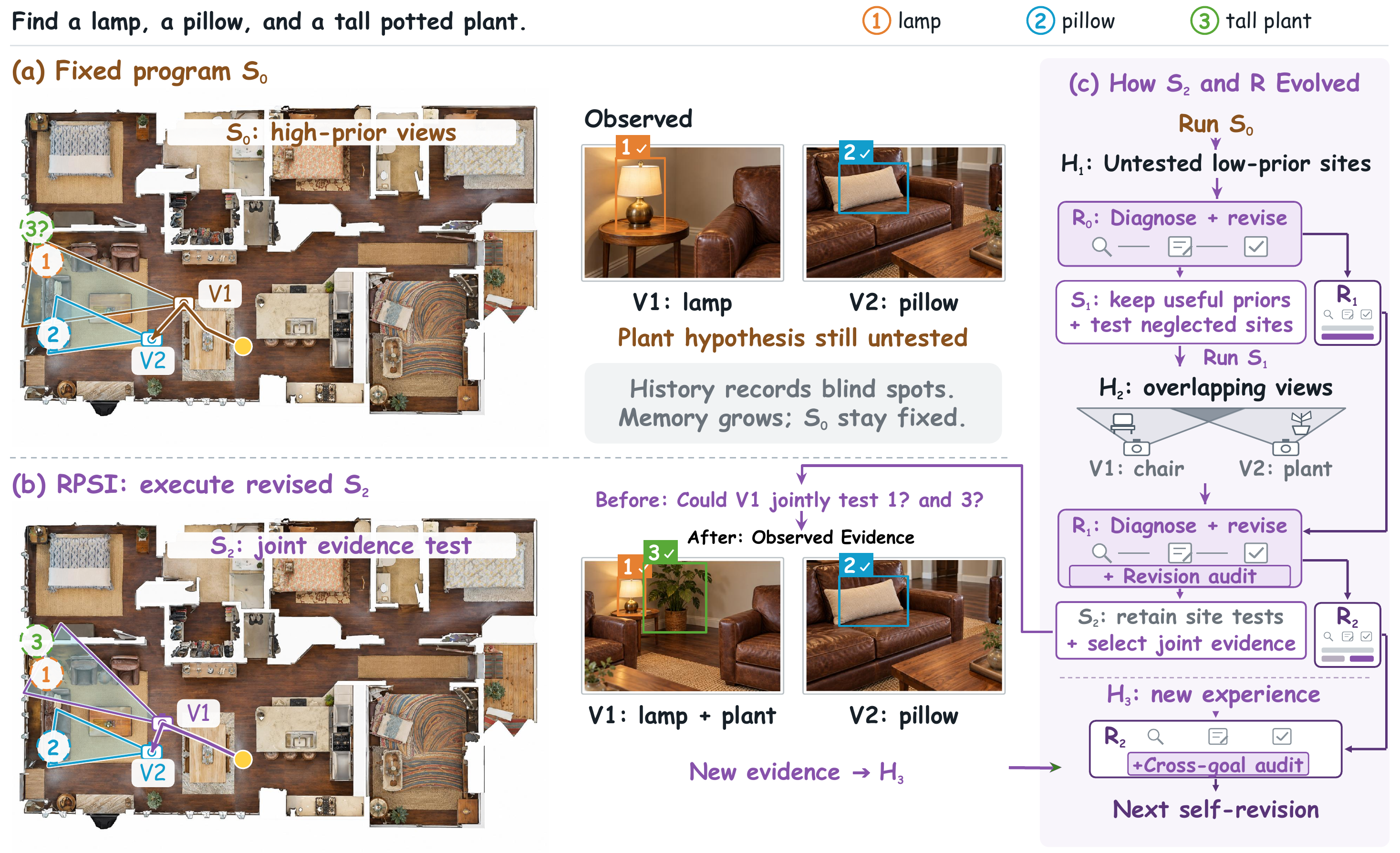}
    \caption{
        \textbf{Recursive improvement across goal searches.}
        \textbf{(a)} $S_0$ finds the lamp and pillow but leaves
        the plant hypothesis untested.
        \textbf{(b)} $S_2$ selects a view that observes
        the lamp and plant.
        \textbf{(c)} Search experience guides navigation revisions,
        while revision outcomes drive the improver's self-revision.
        Accepted improver updates govern subsequent revisions.
        $S_k$ denotes the navigation program $P_k$
        in our method.
    }
    \label{fig:motivation}
\end{figure*}

However, their updates primarily concern navigation rules or
decision guidance, while the procedures that generate these
updates remain predefined. 
In multi-object navigation, subsequent searches can reveal
whether earlier revisions resolved their intended problems.
However, the observations needed to explain an ineffective
revision may not yet be available, and acquiring them consumes
the same movement budget as navigation.
Later searches can therefore expose omissions in the
diagnostic checks or observation requests that guided
earlier repairs.
If this feedback only triggers further policy updates,
the rules producing these omissions may remain unchanged,
even as the navigation policy evolves.
Figure~\ref{fig:motivation} illustrates why improving navigation
also calls for improving the revision process.
An initial revision adds checks of neglected locations to
$S_0$, producing $S_1$. Subsequent searches may still yield
redundant observations, revealing that the repair addressed
where to visit without adequately considering what each view
would reveal. The improver uses this feedback to revise its
own diagnosis and repair rules. The updated improver then revises $S_1$ into $S_2$, which
retains useful site checks and selects views that test
multiple object hypotheses.

To this end, we propose \textbf{RPSI-Nav}, a
\textbf{R}ecursive \textbf{P}olicy \textbf{S}elf-\textbf{I}mprovement
framework for zero-shot open-vocabulary multi-object navigation. Its improvement program diagnoses
search problems, gathers observations that distinguish
competing explanations, and generates targeted revisions
to the navigation program. For each revision, the system
records its intended behavioral change and actual execution
outcome. Subsequent goal searches test these revisions
and provide feedback on the diagnosis and repair decisions
behind them.
When this feedback exposes limitations in its improvement
rules, the current improver applies those rules to propose
revisions to itself. These changes can alter which causes
are considered, which observations are requested, and
which navigation functions are revised. The accepted
version then governs the next round of navigation revisions,
whose outcomes drive further self-revision, closing the
recursive loop. Conditional inheritance carries accepted
updates into later searches where applicable. We evaluate
RPSI-Nav on public open-vocabulary multi-object navigation
benchmarks, complemented by qualitative real-robot experiments.

\section{Related Work}
\label{sec:related_work}

\textbf{Open-vocabulary and multi-object navigation.}
Zero-shot object navigation uses pretrained models for semantic grounding,
commonsense reasoning, and relational search~\cite{gadre2023cow,zhou2023esc,yokoyama2024vlfm,yin2024sgnav}.
OpenFMNav interprets free-form requests~\cite{kuang2024openfmnav}, while
VoroNav combines topology and semantics for exploration~\cite{wu2024voronav}.
MultiON studies semantic memory over object sequences~\cite{wani2020multion};
OneMap and WTRP-Searcher support zero-shot multi-object search through
persistent mapping and global planning~\cite{onemap,wtrp_searcher}.
GOAT-Bench evaluates multimodal lifelong navigation, and 3D-Mem builds reusable
scene memory for exploration and reasoning~\cite{khanna2024goatbench,yang2025threedmem}.
Recent work further advances multimodal goal understanding
and scene representations for navigation~\cite{yin2025unigoal,huang2026msgnav,hu2026astranavmemory,raychaudhuri2025mlfm,ortegapeimbert2025divnav}.
These studies primarily improve environmental representations
and search planning, while we use feedback from successive goal
searches to revise both the navigation program and its
improvement mechanism.

\textbf{Experience-driven navigation improvement.}
Navigation agents increasingly turn execution experience into reusable
rules, memories, and reflective guidance.
Recent methods improve navigation through rule evolution,
memory updates, and reflective guidance~\cite{evolvenav,ge2026evomemnav,li2026metanav,retronav,slow4fast,zhao2026navharness}.
Other approaches improve navigation models through self-reflective
learning and recursive training loops~\cite{lin2025evolvenav,li2026rsinav}.
RPSI-Nav instead studies program-level improvement with frozen foundation models:
subsequent goals provide feedback on earlier revisions, which updates both
the navigation program and the diagnosis, observation-acquisition, and
revision rules that produce future changes.

\textbf{Agents and recursive self-improvement.}
Language feedback and failure analysis support behavioral
revision~\cite{shinn2023reflexion,liu2023reflect,madaan2023selfrefine}.
Programmatic approaches enable reusable skills and automated
optimization of agent workflows and memory mechanisms~\cite{liang2023codeaspolicies,wang2024voyager,hu2025adas,zhang2025aflow,xiong2026alma}.
Recursive self-improvement evolves agent code and its
revision mechanisms~\cite{zelikman2024stop,zhang2026dgm}.
Hyperagents and MetaSkill-Evolve explicitly make the
improvement process itself editable~\cite{hyperagents,wang2026metaskillevolve}.
In robotics, related work improves policy code, execution
harnesses, and diagnostic resources through feedback,
targeted experiments, or selective model updates~\cite{elmaaroufi2026rho,wang2026harnesspai,physevo,song2026embodiedrsi,wang2026adarsi}.
RPSI-Nav studies recursive improvement in continuous visual
search, where evidence for assessing a repair may arrive during
later goals or require additional movement to obtain.
Cross-goal feedback revises how the improver diagnoses failures,
requests observations, and selects navigation changes.
Accepted rules govern subsequent revisions under their
applicable conditions.

\section{Method}
\label{sec:method}

RPSI-Nav leverages sequential goal searches to improve navigation and its revision mechanism. Early searches supply policy-improvement experience, while later searches test whether prior revisions fix their target issues. We link these two feedback streams via recursive self-revision: the improver updates itself, and the updated improver shapes future navigation revisions. Accepted changes transfer across goals under applicable conditions.

\subsection{Task and Program Representation}

In zero-shot open-vocabulary multi-object navigation, a robot
searches for successive language-specified goals in an unknown
environment. Spatial memory can reduce repeated exploration,
but adapting region selection, viewpoints, and revisits requires
changes to search behavior. We therefore maintain executable
programs for both navigation and its improvement, alongside
persistent spatial memory.

Given goals $\mathcal Q=(q^{(1)},\ldots,q^{(J)})$, the robot
uses RGB-D observations $o_t=(I_t,D_t)$ and pose $\xi_t$,
without task-specific training or a prior map.
The sequential protocol reveals each goal after the previous
one is completed; the simultaneous protocol provides all goals
at the start. The observable state is
$z_t=(o_t,\xi_t,q_t,\mathcal M_t,\mathcal H_t,b_t)$,
where $q_t$ is the active issued goal, $\mathcal M_t$ is spatial
memory, $\mathcal H_t$ is interaction history, and $b_t$ is
the remaining movement budget.

At evolution round $k$, the programs are
\begin{equation}
\Theta_k=(P_k,R_k),
\qquad
R_k=(D_k,Q_k,U_k).
\label{eq:programs}
\end{equation}
The navigation program $P_k$ controls location hypotheses,
candidate ranking, viewpoint selection, revisits, and goal
verification. The improvement program $R_k$ contains executable
rules for diagnosis ($D_k$), observation acquisition ($Q_k$),
and revision planning ($U_k$). Its revision targets include
both $P_k$ and itself. The index $k$ is independent of
navigation time and goal indices.

A fixed harness executes the active navigation program:
\begin{equation}
a_t=\mathsf H.\operatorname{Execute}
(P_{\mathrm{act}},z_t;\rho_t),
\label{eq:execute}
\end{equation}
where $P_{\mathrm{act}}$ includes accepted or trial versions,
and $\rho_t$ is an optional observation request.
Search and diagnostic observation share the movement budget.
Diagnosis, coding, and verification agents use separate
contexts and frozen foundation models. $R_k$ supplies their
improvement rules; the harness manages interaction, records,
and version commits.

\subsection{Improving Navigation from Prior Goal}

Earlier goal searches reveal how the robot's decisions affect
search efficiency. For example, a robot searching for a chair
may repeatedly inspect a semantically promising living room
from similar viewpoints. Adding these observations to memory
does not change the ranking or viewpoint rules used for the
next goal. RPSI-Nav converts such experience into reusable
program revisions by tracing search behavior to the rules
that produced it.

The harness records observations, candidate regions and views,
executed choices, and their program origins. $D_k$ uses these
records to identify possible defects, such as unsuitable
region priorities or insufficient viewpoint coverage.
When the causes remain ambiguous, $Q_k$ requests affordable
views that help distinguish them. $U_k$ then selects the
navigation functions and code changes needed for repair.
For target program $X$ and available records $\mathcal B$,
the shared improvement procedure forms
\begin{equation}
\begin{gathered}
\mathcal F^{(0)}=D_k(X,\mathcal B),\quad
\rho=Q_k(X,\mathcal F^{(0)},\mathcal B;z_t),\\
\mathcal B^+=\mathcal B\oplus
\mathsf H.\operatorname{Observe}(\rho;P_{\mathrm{act}},z_t),\\
\Pi_k^X=U_k\!\left(X,D_k(X,\mathcal B^+),\mathcal B^+\right).
\end{gathered}
\label{eq:improver}
\end{equation}
Here, $\mathcal F^{(0)}$ contains candidate explanations,
and $\oplus$ appends actual interaction records.
Without an observation request, $\mathcal B^+=\mathcal B$.
The plan $\Pi_k^X$ specifies the functions, applicability
conditions, and expected behavior of the repair; the coding
agent returns the executable patch
$\mathcal I(X,\mathcal B;R_k)=\mathsf A_{\mathrm{code}}(X,\Pi_k^X)$.

For search experience $\mathcal E_k$ and earlier revision
records $\mathcal A_{<k}$, let
$\mathcal B_k^P=(\mathcal E_k,\mathcal A_{<k})$.
The navigation update is
\begin{equation}
\Delta P_k=\mathcal I(P_k,\mathcal B_k^P;R_k),\quad
\widetilde P_{k+1}=\operatorname{Patch}(P_k,\Delta P_k).
\label{eq:patchP}
\end{equation}

In the chair-search example, suppose $R_0$ attributes repeated
visits to region ranking and generates $P_1$, which lowers
a region's priority after repeated visits without goal
confirmation. This rule can transfer to later searches
under the same conditions. The harness records its expected
effect, whether the patch triggers, how region choices change,
and what new observations result. These records provide
feedback on both the repair and its generating mechanism.

\begin{figure*}[t]
    \centering
    \includegraphics[width=\textwidth]{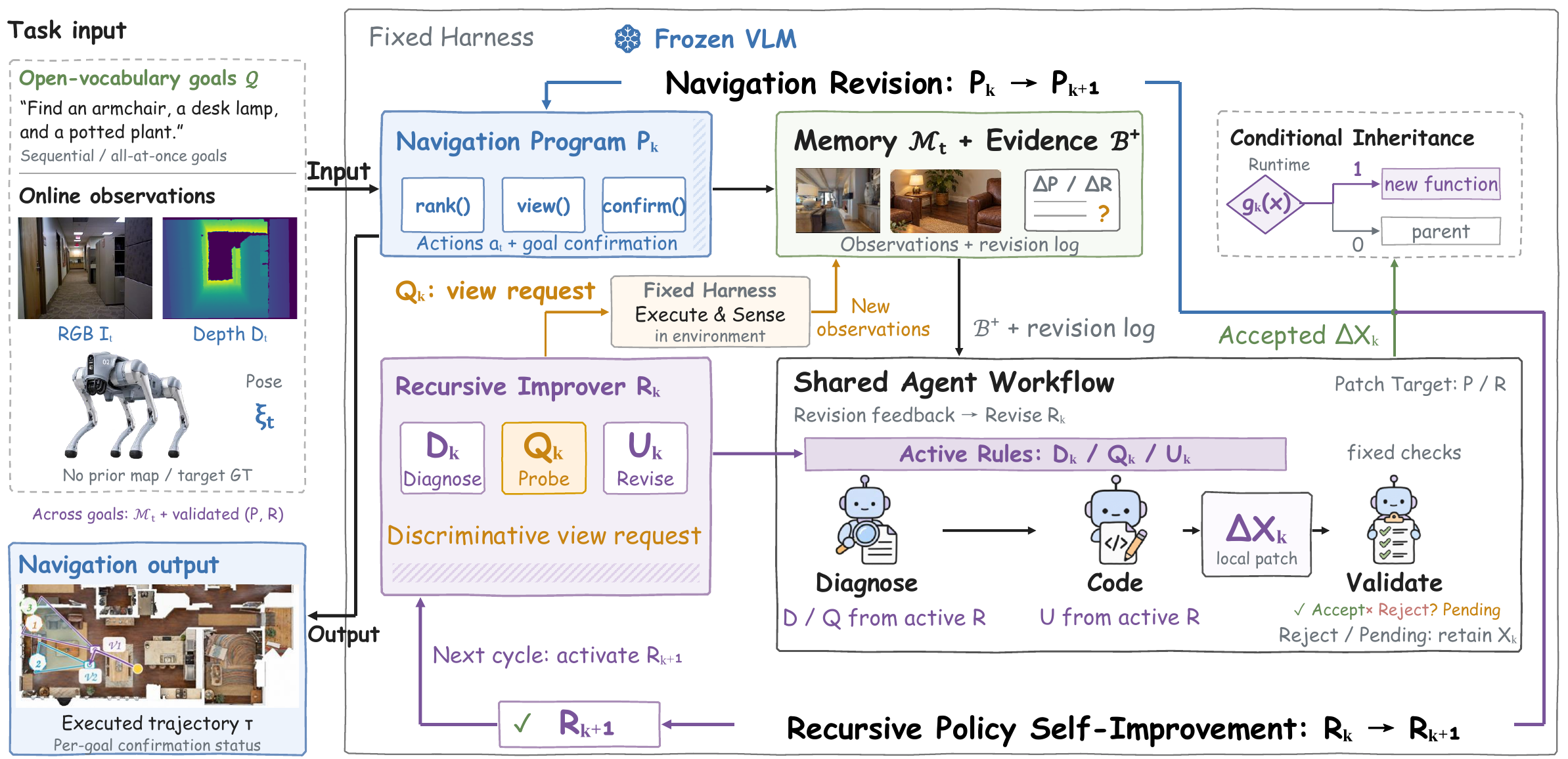}
    \caption{
        \textbf{Overview of RPSI-Nav.}
        Goals $\mathcal Q$, RGB-D observations $(I_t,D_t)$,
        pose $\xi_t$, and memory $\mathcal M_t$ support navigation
        by $P_k$, yielding actions $a_t$, trajectory $\tau$,
        and goal confirmations.
        A fixed harness executes actions with frozen-VLM perception.
        $R_k=(D_k,Q_k,U_k)$ guides diagnosis, observation acquisition,
        and revision through the shared agent workflow.
        Augmented records $\mathcal B^+$ and revision feedback
        support patches $\Delta X_k$, $X\in\{P,R\}$.
        Accepted patches apply under guard $g_k(x)$.
        Revision outcomes drive $R_k$'s self-revision, and accepted
        $R_{k+1}$ governs subsequent navigation revisions,
        closing the recursive loop.
    }
    \label{fig:overview}
\end{figure*}

\subsection{Improving the Improver via Revision Feedback}

In sequential multi-object navigation, experience is selected
by the current search policy, and limitations of a repair may
become visible only during later goal searches.
RPSI-Nav links these observations to earlier patches and the
improvement rules that generated them.
When the records leave competing explanations unresolved,
it requests diagnostic observations within the navigation
budget before revising the improver.
The updated rules then change how subsequent navigation
repairs are diagnosed and selected.

Continuing the example, plant-search records show that $P_1$'s
ranking patch triggers and changes region choices, yet
observations remain redundant. A later side view reveals
previously occluded target cues. This feedback indicates
that the repair addressed where to search without resolving
how to observe the selected regions. Inspecting the revision
trace reveals that $D_0$ selected a ranking explanation
without checking viewpoint coverage.

We associate such feedback with its generating decisions through
program versions, executed rules, and revision identifiers:
\begin{equation}
\mathcal B_k^R=(\mathcal E_k^R,\mathcal A_{\leq k}),
\label{eq:revisionfeedback}
\end{equation}
where $\mathcal A_{\leq k}$ contains revision records available
at the update, and $\mathcal E_k^R$ contains relevant actual
observations, including those from later goals.
The current improver compares expected changes, actual
execution, and search outcomes using its revision-review rules.
The same procedure in Eq.~\eqref{eq:improver} now diagnoses
the rules that generated earlier repairs and plans changes
to those rules:
\begin{equation}
\Delta R_k=\mathcal I(R_k,\mathcal B_k^R;R_k),\;
\widetilde R_{k+1}=\operatorname{Patch}(R_k,\Delta R_k).
\label{eq:patchR}
\end{equation}
Thus, $R_k$ is both the code being revised and the rule set
participating in its revision. Its self-revision rules
are also editable.
In this example, $R_0$ proposes $R_1$ with a
coverage check in $D_1$, a rule in $Q_1$ that requests a
complementary view when coverage is insufficient and the cause
remains ambiguous, and a branch in $U_1$ that selects ranking
or viewpoint repair from the resulting diagnosis.
The change to $Q$ is a reusable condition for future observation
requests, whose execution still depends on the current map
and budget.

The current update uses the accepted snapshot $R_k$.
After verification, accepted $R_{k+1}$ governs the next
navigation revision:
\begin{equation}
\Delta P_{k+1}
=\mathcal I(P_{k+1},\mathcal B_{k+1}^P;R_{k+1}).
\label{eq:nextpatch}
\end{equation}
During a later bed search, $R_1$ can therefore diagnose
repeated doorway observations through coverage checks and
a complementary view. If these support a viewpoint defect,
$U_1$ revises $P_1$'s viewpoint function to favor reachable
views of unobserved candidate locations, producing $P_2$.
The diagnostic observation establishes the repair target;
the revised function changes subsequent search behavior.
Its execution outcomes return to the revision history
and can drive another update of $R_1$.
This closes the recursive loop:
\emph{revision feedback changes the improver's rules,
which change the next navigation revision and generate
new feedback}.

\subsection{Conditional Inheritance Across Goals}
\label{sec:inheritance}

Cross-goal reuse must account for differences in observation
needs. A complementary-view rule is useful when coverage
is insufficient, but need not trigger when the target is
already clearly visible. Each accepted revision therefore
carries an applicability guard:
\begin{equation}
f_{k+1}(x)=
\begin{cases}
\widetilde f_{k+1}(x),&g_k(x)=1,\\
f_k(x),&g_k(x)=0.
\end{cases}
\label{eq:conditional}
\end{equation}
Derived from the revision plan, $g_k$ can use semantic support,
coverage, reachability, and budget. For improver rules,
it can also use the revision target and prior feedback.
Outside the guard, behavior and state writes follow the
parent function; guard evaluation itself does not alter state.

After interface checks, the harness trials each candidate
within $H$ actions from the navigation budget.
Acceptance requires rule activation and observed evidence
of the specified local repair, such as increased coverage
of relevant candidate locations.
For $R$, the revised improvement rules must generate one
navigation patch that passes the same check.
Execution violations or observations contradicting the
repair lead to rejection; insufficient evidence leaves
the update pending.
The resulting status $s_k^X$ determines
\begin{equation}
X_{k+1}=
\begin{cases}
\widetilde X_{k+1},&s_k^X=\mathrm{accept},\\
X_k,&s_k^X\in\{\mathrm{reject},\mathrm{pending}\},
\end{cases}
\quad X\in\{P,R\}.
\label{eq:commit}
\end{equation}
All outcomes enter the revision history.

Within an episode, subsequent goals inherit spatial memory
and accepted programs:
\begin{equation}
\mathcal M_{j+1}^{0}=\mathcal M_j^{\mathrm{end}},\qquad
(P,R)_{j+1}^{0}=(P,R)_j^{\mathrm{accepted}}.
\label{eq:inheritance}
\end{equation}
The inherited programs guide later searches, whose observations
reassess previous repairs and their generating rules.
Independent episodes reset memory and programs according
to the evaluation protocol.

\section{Experiments}
\label{sec:experiments}

We evaluate RPSI-Nav on HM3D multi-object
navigation~\cite{onemap,wtrp_searcher},
HM3D-MFMON~\cite{li2026lifelongcrossnav},
GOAT-Bench~\cite{khanna2024goatbench}.
We build RPSI-Nav upon the publicly released OneMap
framework~\cite{onemap}, which serves as our base navigation
system and primary implementation baseline.
For standard HM3D evaluation, we follow the original
multi-object navigation protocol to examine the benefits
of introducing recursive program self-improvement.
The agent receives RGB-D observations, poses, and issued goals,
with all foundation models frozen.
In standard evaluations, spatial memory and accepted program
updates persist across goals and reset between episodes.
We report SR and SPL, plus PR and PPL for three-goal tasks.
Scores are percentages, and absolute gains are percentage points.

\subsection{Comparative Evaluation}
\label{sec:comparative_evaluation}

\paragraph{Multi-Object Navigation on HM3D}
\label{sec:hm3d_results}
We evaluate RPSI-Nav on the HM3D benchmark following the OneMap protocol~\cite{onemap}.
In the sequential protocol, the next goal is revealed after
the current goal is found, and failure terminates the episode.
We also evaluate the simultaneous protocol of
WTRP-Searcher~\cite{wtrp_searcher}, where all goals are
available from the outset.
As shown in Table~\ref{tab:multi_object}, RPSI-Nav
achieves the best performance across all metrics
under both protocols.
Since RPSI-Nav is built upon OneMap, we first examine
the improvements over this base navigation framework.
Under the sequential protocol, RPSI-Nav improves SR
from 54.20\% to 63.14\% and SPL from 27.80\% to 35.81\%,
yielding gains of \textbf{8.94} and \textbf{8.01}
percentage points, respectively.
These correspond to relative improvements of
16.5\% in SR and 28.8\% in SPL.
Moreover, RPSI-Nav surpasses the stronger WTRP-Searcher
baseline by 5.94 SR and 2.41 SPL points.

\begin{figure*}[t!]
    \centering
    \includegraphics[width=\textwidth]{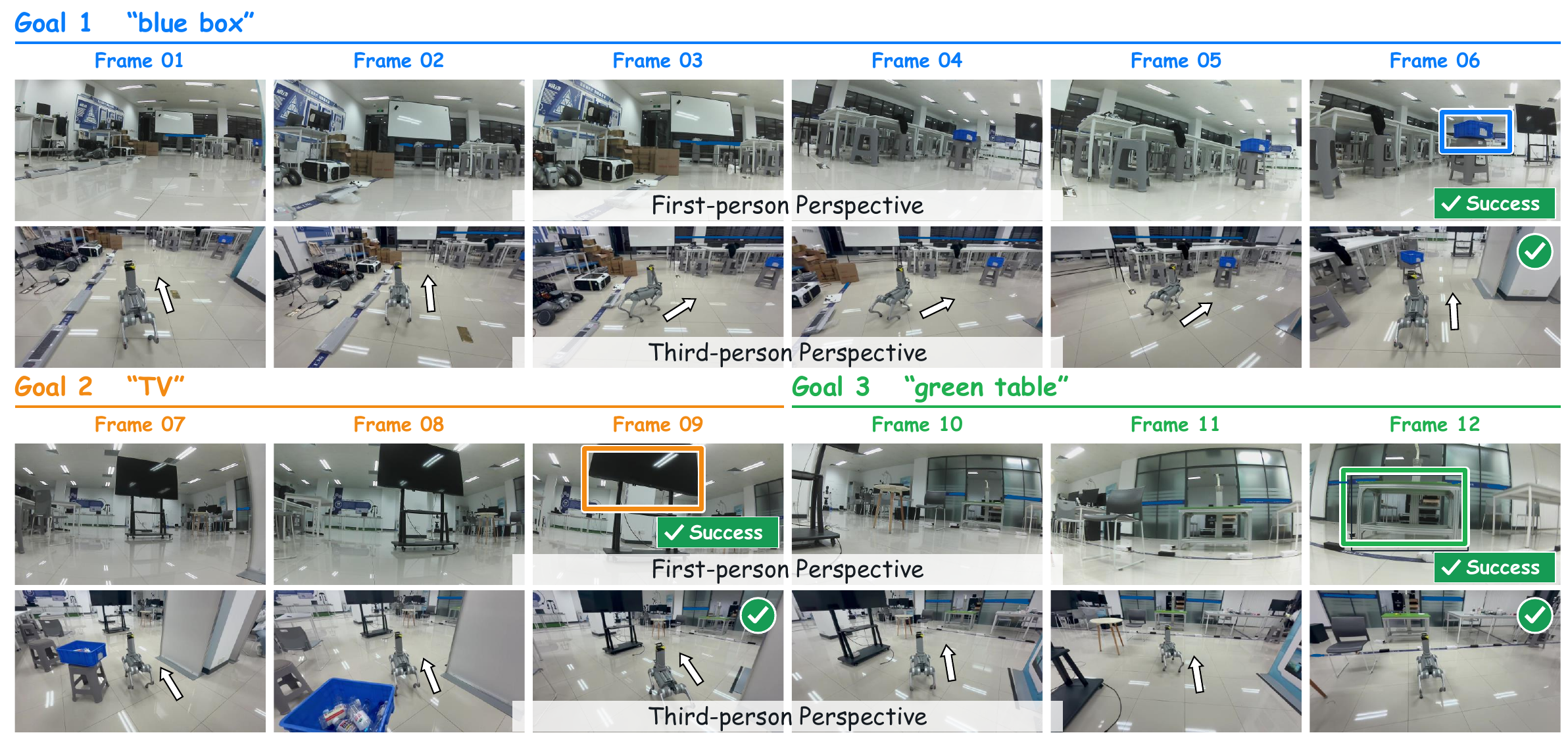}
    \caption{
        \textbf{Real-world multi-object navigation.}
        The robot sequentially searches for a blue box,
        a TV, and a green table.
        First- and third-person views show the search process;
        colored boxes highlight targets, and arrows indicate
        motion directions.
    }
    \label{fig:real_world}
\end{figure*}

Under the simultaneous protocol, RPSI-Nav further
improves upon OneMap by 3.39 SR and 2.78 SPL points,
while exceeding WTRP-Searcher by 2.53 SR and
3.41 SPL points.
These results demonstrate consistent improvements
over the underlying navigation framework and
competitive performance against stronger baselines.
We deploy RPSI-Nav on a quadruped robot in an indoor
environment.
As shown in Fig.~\ref{fig:real_world}, the robot sequentially
locates a blue box, a TV, and a green table while navigating
around furniture.

\begin{table}[t!]
    \centering
    \caption{Multi-object navigation on HM3D.}
    \label{tab:multi_object}
    \small
    \setlength{\tabcolsep}{3.5pt}
    \renewcommand{\arraystretch}{1.15}
    \resizebox{\columnwidth}{!}{%
    \begin{tabular}{lccccc}
        \toprule
        \multirow{2}{*}{Method}
        & \multirow{2}{*}{Type}
        & \multicolumn{4}{c}{HM3D} \\
        \cmidrule(lr){3-6}
        & & SR$\uparrow$ & SPL$\uparrow$
          & PR$\uparrow$ & PPL$\uparrow$ \\
        \midrule
        VLFM~\cite{yokoyama2024vlfm}
        & \multirow{4}{*}{Sequential}
        & 44.92 & 19.90 & 57.34 & 24.87 \\
        OneMap~\cite{onemap}
        & & 54.20 & 27.80 & 65.50 & 32.70 \\
        WTRP~\cite{wtrp_searcher}
        & & 57.20 & 33.40 & 67.10 & 38.20 \\
        \textbf{RPSI-Nav (Ours)}
        & & \textbf{63.14} & \textbf{35.81}
          & \textbf{69.49} & \textbf{40.07} \\
        \midrule
        OneMap~\cite{onemap}
        & \multirow{3}{*}{Simultaneous}
        & 52.54 & 23.63 & 68.93 & 31.40 \\
        WTRP~\cite{wtrp_searcher}
        & & 53.40 & 23.00 & 68.90 & 28.90 \\
        \textbf{RPSI-Nav (Ours)}
        & & \textbf{55.93} & \textbf{26.41}
          & \textbf{70.06} & \textbf{32.74} \\
        \bottomrule
    \end{tabular}%
    }
\end{table}

\textbf{Multi-Floor Navigation. }
\label{sec:multifloor_results}
HM3D-MFMON~\cite{li2026lifelongcrossnav} contains 927
three-goal episodes across 36 multi-floor scenes.
Its 288-episode Cross-Floor-Required subset specifically
tests sequences requiring at least one floor transition.
We follow the benchmark's sequential protocol and post-hoc
stage-wise shortest-path evaluation.
On the full set, RPSI-Nav achieves 39.81 SR and 15.41 SPL,
improving over LifelongCrossNav by 3.24 and 3.03 points
(Table~\ref{tab:mfmon_all}).
On the Cross-Floor-Required subset, it reaches 15.97 SR
and 4.87 SPL, with gains of 3.47 and 1.76 points
(Table~\ref{tab:mfmon_cross_floor}).
Together with higher PR and PPL, these results show improved
sequence completion and search efficiency in environments
requiring floor transitions.

\begin{table}[t]
    \centering
    \caption{Multi-object navigation on HM3D-MFMON.}
    \label{tab:mfmon_all}
    \small
    \setlength{\tabcolsep}{3pt}
    \renewcommand{\arraystretch}{1.1}
    \begin{tabular*}{\columnwidth}{
        @{\extracolsep{\fill}}lcccc@{}
    }
        \toprule
        Method & SR$\uparrow$ & SPL$\uparrow$
               & PR$\uparrow$ & PPL$\uparrow$ \\
        \midrule
        OneMap~\cite{onemap}
            & 16.83 & 6.79 & 33.69 & 13.95 \\
        LifelongCrossNav~\cite{li2026lifelongcrossnav}
            & 36.57 & 12.38 & 54.33 & 18.86 \\
        \midrule
        \textbf{RPSI-Nav (Ours)}
            & \textbf{39.81} & \textbf{15.41}
            & \textbf{56.71} & \textbf{21.49} \\
        \bottomrule
    \end{tabular*}
\end{table}

\begin{table}[t]
    \centering
    \caption{Multi-object navigation on the 288
    Cross-Floor-Required HM3D-MFMON episodes.}
    \label{tab:mfmon_cross_floor}
    \small
    \setlength{\tabcolsep}{3pt}
    \renewcommand{\arraystretch}{1.1}
    \begin{tabular*}{\columnwidth}{
        @{\extracolsep{\fill}}lcccc@{}
    }
        \toprule
        Method & SR$\uparrow$ & SPL$\uparrow$
               & PR$\uparrow$ & PPL$\uparrow$ \\
        \midrule
        OneMap~\cite{onemap}
            & 0.00 & 0.00 & 18.52 & 7.97 \\
        LifelongCrossNav~\cite{li2026lifelongcrossnav}
            & 12.50 & 3.11 & 33.68 & 10.86 \\
        \midrule
        \textbf{RPSI-Nav (Ours)}
            & \textbf{15.97} & \textbf{4.87}
            & \textbf{34.95} & \textbf{12.31} \\
        \bottomrule
    \end{tabular*}
\end{table}

\textbf{Multimodal Navigation on GOAT-Bench. }
\label{sec:goat_results}
GOAT-Bench~\cite{khanna2024goatbench} specifies successive
goals through category names, language descriptions,
and reference images.
Following 3D-Mem~\cite{yang2025threedmem}, we evaluate one
episode per Val Unseen scene, covering 278 subtasks across
36 scenes, and report subtask SR and SPL.
We compare with the baselines summarized by
SCOPE~\cite{wang2026scope} and additionally include
HIMM~\cite{li2026himm}.
Table~\ref{tab:goatbench} shows that RPSI-Nav achieves
the highest SR of 75.2 and the second-highest SPL of 54.7.
It exceeds SCOPE by 1.5 SR and 1.2 SPL points,
while HIMM retains the highest SPL.
These results demonstrate effective sequential search
across different goal modalities.

\begin{table}[t]
    \centering
    \caption{Comparison on GOAT-Bench.}
    \label{tab:goatbench}
    \small
    \renewcommand{\arraystretch}{1.1}
    \begin{tabular*}{\columnwidth}{
        @{\extracolsep{\fill}}lcc@{}
    }
        \toprule
        Method & SR$\uparrow$ & SPL$\uparrow$ \\
        \midrule
        Explore-EQA~\cite{ren2024exploreeqa}
            & 55.0 & 37.9 \\
        CG + Frontier~\cite{gu2024conceptgraphs}
            & 61.5 & 41.3 \\
        3D-Mem~\cite{yang2025threedmem}
            & 69.1 & 48.9 \\
        HIMM~\cite{li2026himm}
            & 72.8 & \textbf{56.1} \\
        SCOPE~\cite{wang2026scope}
            & 73.7 & 53.5 \\
        \midrule
        \textbf{RPSI-Nav (Ours)}
            & \textbf{75.2} & 54.7 \\
        \bottomrule
    \end{tabular*}
\end{table}

\subsection{Ablation Study}
\label{sec:rsi_experiments}

\paragraph{Effect of improver self-revision.}
We compare four variants: no program revision, a fixed
improver $R_0$, recursive improvement with fixed acquisition
rules $Q_0$, and full RPSI-Nav.
The fixed-$Q_0$ variant allows $P$, $D$, and $U$ to evolve,
while $Q_0$ continues to use current diagnoses, maps, and
budgets without changing its rules.
Table~\ref{tab:rsi_ablation} shows that revising $P$ with
fixed $R_0$ improves SR and SPL by 3.5 and 2.6 percentage
points over no revision.
Allowing $D/U$ to evolve with $Q_0$ fixed adds 2.1 and
1.7 points, respectively.
Full RPSI-Nav further improves SR and SPL by 3.3 and
3.7 points over fixed $Q_0$, supporting the value of
evolving observation-acquisition rules within the
recursive improvement process.

\begin{table}[t]
    \centering
    \caption{Effect of improver self-revision.
    Scores are means over independent runs (\%).}
    \label{tab:rsi_ablation}
    \small
    \setlength{\tabcolsep}{3pt}
    \renewcommand{\arraystretch}{1.1}
    \begin{tabular*}{\columnwidth}{
        @{\extracolsep{\fill}}ccccc@{}
    }
        \toprule
        Variant
        & SR$\uparrow$
        & SPL$\uparrow$
        & PR$\uparrow$
        & PPL$\uparrow$ \\
        \midrule
No program revision
    & 54.2 & 27.8 & 65.5 & 32.7 \\
Fixed improver $R_0$
    & 57.7 & 30.4 & 68.4 & 36.4 \\
Recursive, fixed $Q_0$
    & 59.8 & 32.1 & \textbf{70.1} & 38.2 \\
\textbf{RPSI-Nav}
    & \textbf{63.1} & \textbf{35.8}
    & 69.5 & \textbf{40.1} \\
        \bottomrule
    \end{tabular*}
\end{table}

\textbf{Effect of cross-goal improver inheritance. }
To examine whether evolved improvement rules benefit subsequent goals,
we compare RPSI-Nav with a variant that resets $R$ to
$R_0$ at each goal transition.
The accepted navigation program, spatial memory, and revision
history are retained across goals, and both $P$ and $R$ remain editable
within each goal search.
We evaluate both variants across five paired runs on the 236 sequential three-goal episodes.

Retaining accepted $R$ updates across goals in full RPSI-Nav
yields a further gain of 2.2 SR and 2.0 SPL points over
resetting $R$ at goal transitions
(Table~\ref{tab:cross_goal_inheritance}).
The reset variant still improves over the fixed-$R_0$ baseline
in Table~\ref{tab:rsi_ablation} by 3.2 SR and 3.4 SPL percentage
points (from 57.7/30.4 to 60.9/33.8), reflecting within-goal
improver self-revision.

\begin{table}[t]
    \centering
    \caption{
        Effect of cross-goal improver inheritance on HM3D.
        Scores are means over five independent runs (\%).
    }
    \label{tab:cross_goal_inheritance}
    \small
    \setlength{\tabcolsep}{3pt}
    \renewcommand{\arraystretch}{1.1}
    \begin{tabular*}{\columnwidth}{
        @{\extracolsep{\fill}}lcccc@{}
    }
        \toprule
        Variant
        & SR$\uparrow$
        & SPL$\uparrow$
        & PR$\uparrow$
        & PPL$\uparrow$ \\
        \midrule
        Reset $R$ across goals
            & 60.9 & 33.8 & 68.8 & 38.7 \\
        \textbf{RPSI-Nav}
            & \textbf{63.1} & \textbf{35.8} & \textbf{69.5} & \textbf{40.1} \\
        \bottomrule
    \end{tabular*}
\end{table}

\textbf{Does the improver improve? }
To assess improvement ability directly, we compare improver
snapshots from different evolution stages at matched
navigation starting points.
We partition scenes into disjoint evolution, revision, and
test sets in a $2{:}1{:}1$ ratio.
During 100 three-goal evolution episodes, accepted programs
and revision records persist, while spatial memory resets
between episodes.
We save $R^{[n]}$ after $n\in\{0,50,100\}$ episodes.
Each snapshot remains fixed while revising two reference
programs: the initial $P_0$ and $P_{\mathrm{ref}}$, obtained
after 50 evolution episodes with fixed $R_0$.
For each reference, all snapshots independently revise ten matched cases from the revision scenes; rejected patches retain the reference program.
Resulting programs $P^{(R^{[n]})}$ are evaluated on the same
100 test episodes with program updates disabled and spatial
memory reset per episode, measuring the gain
$\Delta M = M(P^{(R^{[n]})})-M(P)$ for $M\in\{\mathrm{SR},\mathrm{SPL}\}$.
Table~\ref{tab:improver_quality} shows that the evolved improver
generates more effective revisions for both initial and
previously improved navigation programs.
Averaged over cases, $R^{[100]}$ produces an additional
2.5-point SR and 2.2-point SPL gain from $P_0$, and 1.9-point
SR and 1.7-point SPL gain from $P_{\mathrm{ref}}$,
compared with $R^{[0]}$.

\begin{table}[t]
    \centering
    \caption{Mean navigation gains produced by improver
    snapshots, relative to each reference program.
    }
    \label{tab:improver_quality}
    \small
    \setlength{\tabcolsep}{5pt}
    \renewcommand{\arraystretch}{1.1}
    \begin{tabular*}{\columnwidth}{
        @{\extracolsep{\fill}}cccc@{}
    }
        \toprule
        Reference & Improver
        & $\Delta$SR$\uparrow$
        & $\Delta$SPL$\uparrow$ \\
        \midrule
        $P_0$ & $R^{[0]}$
            & +2.2 & +1.5 \\
              & $R^{[50]}$
            & +3.1 & +2.3 \\
              & $R^{[100]}$
            & \textbf{+4.7} & \textbf{+3.7} \\
        \midrule
        $P_{\mathrm{ref}}$ & $R^{[0]}$
            & +0.7 & +0.5 \\
                          & $R^{[50]}$
            & +1.7 & +1.4 \\
                          & $R^{[100]}$
            & \textbf{+2.6} & \textbf{+2.2} \\
        \bottomrule
    \end{tabular*}
\end{table}

\textbf{Self-revision from an evolved improver. }
To examine whether self-revision remains useful beyond the initial improver,
we compare two variants initialized with the same navigation program $P_{\mathrm{ref}}$
and evolved improver $R^\ast=R^{[100]}$.
The frozen variant updates only $P$, whereas the adaptive variant updates both $P$ and $R$.
Both receive the same interaction and total revision budgets.

We evaluate 100 three-goal episodes held out from improver evolution and checkpoint selection over five paired runs $i\in\{1,\ldots,5\}$, with identical episode orders and initializations.
Each episode begins with the same reference programs and empty spatial memory; accepted updates persist across goals within an episode.
We report marginal scores across runs and the paired run-level difference
$d_i=M_i(\text{Adaptive }R^\ast)-M_i(\text{Frozen }R^\ast)$.
As shown in Table~\ref{tab:evolved_improver_control}, revising $R^\ast$ yields a paired gain of $+1.8 \pm 0.45$ SR points and $+1.2 \pm 0.25$ SPL over freezing it. Self-revision remains beneficial even with a pre-evolved improver.

\begin{table}[t]
    \centering
    \caption{
        Frozen versus adaptive evolved improvers across five paired runs (100 episodes per run).
    }
    \label{tab:evolved_improver_control}
    \small
    \setlength{\tabcolsep}{4pt}
    \begin{tabular*}{\columnwidth}{
        @{\extracolsep{\fill}}lcc@{}
    }
        \toprule
        Variant & SR$\uparrow$ & SPL$\uparrow$ \\
        \midrule
        Frozen $R^\ast$
            & $62.4 \pm 1.52$
            & $35.0 \pm 0.89$ \\
        Adaptive $R^\ast$
            & $64.2 \pm 1.48$
            & $36.2 \pm 0.87$ \\
        \midrule
        Paired gain ($\Delta$)
            & $\mathbf{+1.8 \pm 0.45}$
            & $\mathbf{+1.2 \pm 0.25}$ \\
        \bottomrule
    \end{tabular*}
\end{table}

\begin{figure*}[t!]
    \centering
    \includegraphics[width=\textwidth]{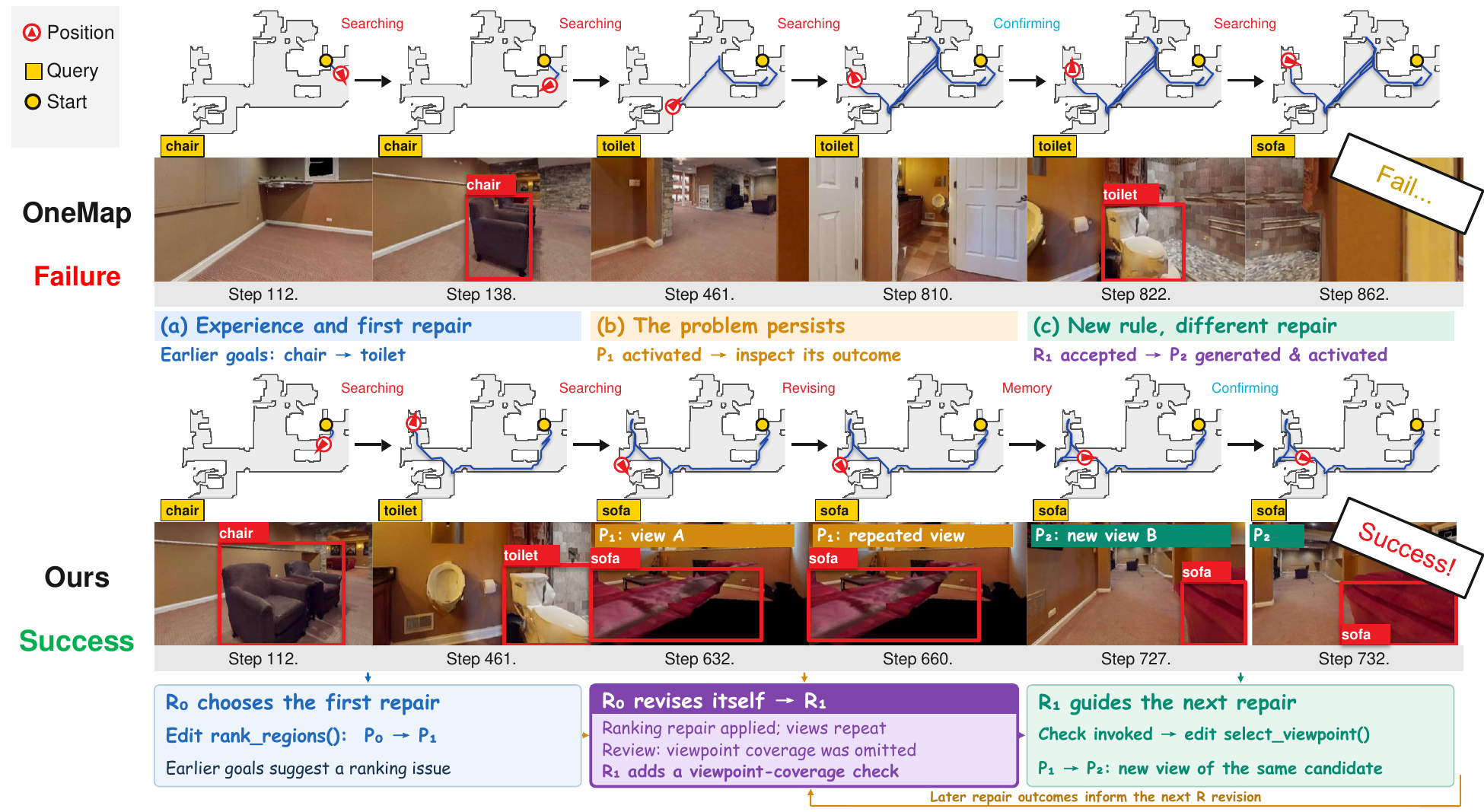}
    \caption{
        \textbf{Navigation comparison and a representative RSI trace.}
        \textbf{Upper row:} OneMap fails to complete the goal sequence.
        \textbf{Lower row:} RPSI-Nav completes the chair, toilet,
        and sofa searches.
        Top-down maps and RGB views show selected timesteps;
        boxes annotate visible targets.
        The three stages illustrate the recursive repair process:
        \textbf{(a)} $R_0$ revises region ranking, producing $P_1$.
        \textbf{(b)} Repeated views prompt $R_0$ to review
        the repair and add a viewpoint-coverage check,
        yielding $R_1$.
        \textbf{(c)} $R_1$ invokes this check and redirects
        the next repair to viewpoint selection, producing $P_2$.
        $P_k$ and $R_k$ denote the navigation and improvement
        programs, respectively.
    }
    \label{fig:rsi_case}
\end{figure*}

\textbf{Recursive execution within episodes. }
Table~\ref{tab:recursive_execution} reports execution statistics for sequential episodes in this setting, including early failures.
Cross-goal revision feedback occurs when observations from a later goal
are linked to review a patch or rule from an earlier goal (67/236 episodes).
A completed recursive chain requires an accepted improver's
revised rules to subsequently generate a new, accepted navigation patch
whose modified behavior executes before episode termination, excluding
temporary validation probes during the improver's own trial.
The chain completes in 84 episodes (35.59\%): 58 episodes (24.58\%)
involve cross-goal recursion where an $R$ update accepted in goal $j$
generates and executes a new $P$ patch in a later goal $j'>j$, while
the remaining 26 episodes (11.02\%) complete the chain exclusively
within a single goal search.

\begin{table}[t]
    \centering
    \caption{Recursive execution statistics across all 236 sequential three-goal episodes.}
    \label{tab:recursive_execution}
    \small
    \setlength{\tabcolsep}{3pt}
    \renewcommand{\arraystretch}{1.1}
    \begin{tabular*}{\columnwidth}{@{\extracolsep{\fill}} l c c @{}}
        \toprule
        Event & Episodes & Rate (\%) \\
        \midrule
        Accepted $R$ update               & 120/236 & 50.85 \\
        Cross-goal revision feedback      & 67/236  & 28.39 \\
        Completed recursive chain (total) & 84/236  & 35.59 \\
        \quad Cross-goal                  & 58/236  & 24.58 \\
        \quad Within-goal only            & 26/236  & 11.02 \\
        \bottomrule
    \end{tabular*}
\end{table}

\textbf{RSI Case Study in Simulation. }
Figure~\ref{fig:rsi_case} compares RPSI-Nav with
OneMap~\cite{onemap} in the same scene with the same goal
sequence. RPSI-Nav completes the successive searches for
a chair, a toilet, and a sofa, whereas OneMap does not
complete the sequence.
The execution trace in the figure shows how revision feedback changes subsequent repairs.
Initially, $R_0$ revises region ranking, producing $P_1$.
When similar views persist during the sofa search,
$R_0$ reviews the earlier repair and identifies an omitted
viewpoint-coverage check. It adds this check to its own
diagnosis rules, yielding $R_1$.
The updated improver distinguishes poor region selection
from redundant observations within a region and directs
the next repair to \texttt{select\_viewpoint()}, producing
$P_2$ and a different view of the same candidate.
Here, revision feedback changes the improver's diagnostic
rule, which then changes the next navigation repair.

\section{Conclusion}
We presented RPSI-Nav, a recursive self-improvement framework for zero-shot open-vocabulary multi-object navigation. Using feedback from sequential searches, RPSI-Nav revises both navigation programs and its own diagnostic, observation-acquisition, and revision rules, transferring verified updates across goals under applicability guards. Benchmark evaluations and real-robot deployments demonstrate its effectiveness and practical feasibility.

\section*{LLM Usage Statement}

ChatGPT was used to assist with research discussions,
literature organization, manuscript drafting, and language
refinement. OpenAI Codex was used to assist with figure
preparation and refinement.
As part of the proposed RPSI-Nav framework,
\texttt{Qwen2.5-Coder-7B-Instruct} serves as a frozen language
model for diagnosis, code generation, and verification-related
reasoning. The authors take full responsibility for the scientific content,
implementation, experimental results, references, and final
presentation of this work.

{\small\bibliographystyle{ieeenat_fullname}\bibliography{references}}
\end{document}